\documentclass{article}
\usepackage{stywhispers,amsmath,epsfig}

\title{An Interpretable Neural Network for Vegetation Phenotyping with Visualization of Trait-Based Spectral Features}

\name{William Basener, Abigail Basener, Michael Luegering\thanks{Funding by USACE-ERDC Engineering With Nature}\thanks{This material is based upon work supported by the National Science Foundation under Grant No 2319470}}
\address{School of Data Science, University of Virginia\\
College of Computer, Mathematical, and Natural Sciences, University of Marlyand\\
School of Architecture, University of Virginia}
\begin{document}
%
\maketitle
\begin{abstract}
Plant phenotyping is the assessment of a plant's traits and plant identification is the process of determining the category such as genus and species. In this paper we present an interpretable neural network trained on the UPWINS spectral library which contains spectra with rich metadata across variation in species, health, growth stage, annual variation, and environmental conditions for 13 selected indicator species and natural common background species.  We show that the neurons in the network learn spectral indicators for chemical and physiological traits through visualization of the network weights, and we show how these traits are combined by the network for species identification with an accuracy around 90\% on a test set.  While neural networks are often perceived as `black box' classifiers, our work shows that they can be in fact more explainable and informative than other machine learning methods.  We show that the neurons learn fundamental traits about the vegetation, for example the composition of different types of chlorophyll present which indicates species as well as response to illumination conditions.  There is clear excess training capacity in our network, and we expect that as the UPWINS spectral library continues to grow the approach in this paper will provide further foundational insights in  understanding plant traits. This provides a methodology for designing and interpreting neural networks on spectral data in general, and provides a framework for using neural networks with hyperspectral imagery for understanding vegetation that is extendable to other domains.
\end{abstract}
\section{Introduction}
\label{sec:intro}
Hyperspectral and multispectral imagery provide a spectrum for each pixel that contains information about the materials present.  Multispectral imagery has been used in a remote sensing context to quantify and understand vegetation since the 1970s~\cite{MultispectralVegetation1980}. Hyperspectral imagery over the VNIRSWIR spectral range (400nm to 2500nm) provides a reflectance spectrum for each pixel. Each value in this spectrum is a measurement of light in a narrow range of wavelengths. Because the fraction of light reflected by a material depends on the relationship between molecular bond structures and the wavelength-dependent energy of the light, hyperspectral pixel spectra contain information about the molecular composition of the materials in the image.  

In this paper we use a growing complex highly-documented spectral library from the UPWINS project for training and testing of algorithms for phenotyping and species species identification with over 1,000 field spectrometer measurements.  We test 26 machine learning algorithms for predicting vegetation species and the full species-health-growth stage information. The library does not yet contain sufficient samples of all categories to make conclusions regarding prediction on the full categorization, but we show that Linear Discriminant Analysis (LDA) and a simple but properly trained neural network (NN) both provide accuracy of around 90\% in predictions across over 13 species. Analysis of weights in the NN show that it is learning chemically meaningful information, for example relative quantities of different types of chlorophyll and other pigments. This suggests that the model is learning meaningful information, rather than overtraining, which suggests that the model, as well as the general methodology, may be robust to broader application. 

The UPWINS (Urban Planning With Integrated Natural Systems) project is continuing at least through 2026, and we will continue collecting ASD fieldspectrometer spectra from vegetation species in natural and intentionally modified environments, with documented variation in soil type, in nutrient, water and salinity content as well as temperature variation and other factors. We will also be collecting hyperspectral imagery (Headwall, Specim) and LiDAR (Velodyne) data over natural areas with our target species, from both UAVs and aircraft, and will be  multispectral satellite (MAXAR Worldview) and SAR (Umbra) imagery. All data is being shared openly, to the extent possible, to support our research and application communities.  The UPWINS project is directed toward developing methodologies for urban planning that integrates with, rather than attempts to constrain, natural systems.  An essential component of this approach to urban planning is assesing and monitoring ecological, geological, and hydrological systems at scale, including phenotyping of vegetation as an indicator of local environment. Remote sensing data collection along with substantial documented fieldspectrometry data, along with development and sharing of algorithms for environmental assessment from this data, is a major portion of this effort.  We hope that sharing of data and results will benefit research in remote sensing and machine learning algorithm development for these and related objectives.

\subsection{Background}\label{ss:background}
In this subsection, we present progress and limitations in use of machine learning on spectral imagery for understanding of vegetation. We use the term machine learning (ML) broadly, to include methods for regression, classification and related tasks for processing of data, recognizing that most or all of these methods can be understood via statistics.  We use deep learning (DL) to refer to the ML algorithms that employ a neural network with many layers. Works discussed in this subsection might use the terms with a different emphasis, but the meaning should be understandable from the context to the extent that it is important.

Hyperspectral imagery has potential for fast, systemic, phenotyping of plants. Plant phenotyping is the analysis of plant structural, chemical, and functional traits.  This is important for selecting plants in crop breeding to enhance desirable traits that increase resiliency and yield.  Phenotyping is a bottleneck in the crop breeding process~\cite{furbank2011phenomics,Pieruschka2023}. Specifically, in 2023 Pieruschka et al state that
\begin{quote}
the analysis of crop performance with respect to structure (root, shoot architecture, leaf angle, etc.), function (photosynthesis, transpiration, growth etc.), quality (chemical composition) and interaction with the environment i.e. phenotyping remains still one of the largest bottlenecks in basic and applied plant sciences.
\end{quote}
Bibliometric analysis of research publications in plant phenotyping show that there was a notable increase in publications on phenotyping using machine learning and deep learning methods with sensor data began in 2016~\cite{Kaur2022}, corresponding to advancements in both sensors and algorithms. However, as noted in~\cite{Pieruschka2023} and elsewhere, there is still significant need for improvement.

Processing of spectral data using Machine Learning (ML) and Deep Learning (DL) algorithms for plant phenotyping has shown benefits for specific crop varieties in controlled environments. However, adoption of these methods broadly is limited by a number of practical factors.  A research review by Shuai et al. in 2024~\cite{Shuai2024} on the use of DL with hyperspectral imaging for agriculture concluded that while hyperspectral imagery capture important information about vegetation, a major limitation is the ``limited labeled samples, homospectral or isospectral characteristics''.  There is an insufficient number of publicly available labeled.  The spectra that are available do not cover the variation in vegetation species, growth stage, health, environmental conditions, disease states, and other factors that are required for prediction.  Compounding this is the fact that DL algorithms require very large quantities of training data. This study also cites the intrinsic ``high data dimensionality'' of hyperspectral imagery as further compounding the limitations from training data that lacks quantity, specificity in labeling, statistical variance, and variance in vegetation phenotype. The study suggests that band selection has potential to reduce problems from the dimensionality and correlation, but that currently methods for``the selection of effective spectral bands'' are insufficient and ``extracting effective feature information from raw HSI data remains a challenge.'' The study concluded that ``agricultural HSI analysis still faces challenges and research gaps in relation to DL's reliance on large-scale high-quality data, limited model interpretability, and the complexities of training intricate models''~\cite{Shuai2024}. This is not to say that either ML or DL methods, or data from hyperspectral sensors, are insufficient for broad practical agricultural and general vegetation analysis, but that understanding the specific current limitations and challenges is important for guiding current and future research in sensor development, algorithm development, training data collection, and implementation protocols.

A systematic literature survey on imaging sensors and artificial intelligence (including ML and DL) for plant stress research with imaging~\cite{Walsh2024} selected 2,704 published papers based on keywords that were then distilled down to 262 papers for in depth analysis. This study also identified open data sets for training and validation as a primary catalyst for advancements in areas where such datasets have been generated, and critical limitation in areas where datasets are not available.  The analysis observes that Kaggle.com and Data.gov are effective platforms for datset sharing and algorithm testing, and cites a few particularly impactful datasets of visual imagery of stressed plants, with the most impactful being the PlanetVilliage dataset ~\cite{hughes2015open} with 50,000 images.  However, no significant datasets were cited supporting hyperspectral imagery. Of the 145 papers in the analysis using deep learning, only 5 involved hyperspectral imagery.  They conclude that 

\begin{quote}
We theorize that spectral imaging requires complex models to identify features (e.g., chlorophyll content), and ML models simplify the interpretation of the results, providing clear rules for decision-making. In contrast, DL models, known for their black-box nature, hinder the biological meaning behind their predictions. The limited availability of spectral imaging datasets, specifically open-source, is another reason behind the low use of DL models, which require large labeled datasets to learn and make accurate predictions.
\end{quote}

The analysis showed that most research focused on biotic stress (stress from biological sources), with only 5\% of research studying abiotic stress (stress from inorganic factors, such as extreme temperatures, drought and flooding, salinity, nutrient deficiency, and metal toxicity). While biotic stress is observable in visual imagery in available datasets, abiotic stress is often less obvious but observable in alteration to spectral values.

Research reviews on use of UAVa for phenotyping of vegetation indicate that most or nearly all studies involve crops rather than in situ natural vegetation and the limited reliability and consistency of collection and processing algorithms limit their use~\cite{Bongomin2024}.  A study of published research from 2019-2024 on crop phenotyping using imagery from UAVs~\cite{Tanaka2024} documented the potential of hyperspectral imagery to measure chlorophyll and N compounds, but concluded that the higher model complexity hinders model performance, and improvement in prediction performance from complex models is not a significant improvement over simpler models. 

Overall, these research studies indicate that a lack in training data, especially adequately labeled data that is sufficient in quantity and variation, is a roadblock to advancing machine learning and deep learning for vegetation analysis and plant phenotyping.  Moreover, they suggest that interpretable models that are robust (i.e. low variance models) are preferable when possible. Such models will generalize to new environments and tasks. They will also be easier to retrain, and be less susceptible to overtraining with datasets that are limited in size or variation.  The combination of interpretable models and sufficient data for training and evaluation will also provide inference into vegetation chemistry and function using model parameters.

\section{Data}
\begin{figure}[htb]
\begin{minipage}[b]{1.0\linewidth}
  \centering
	 \centerline{\epsfig{figure=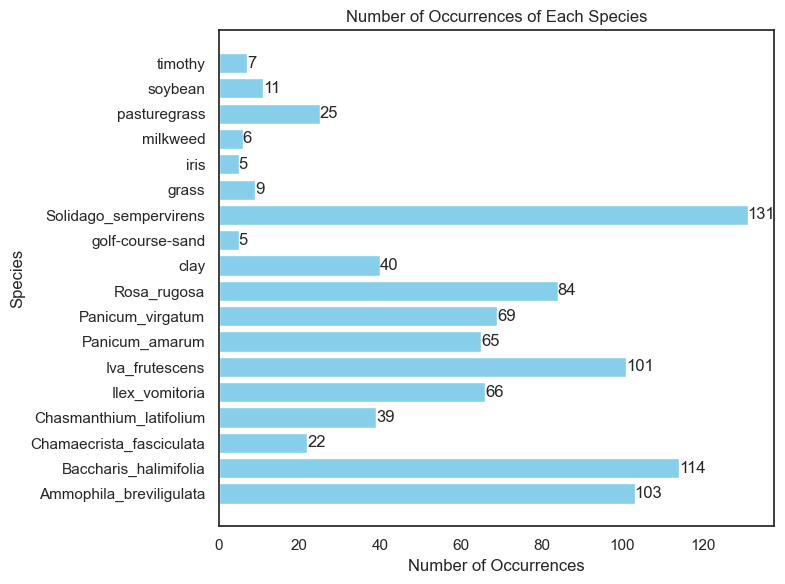, width=8.5cm}}
\end{minipage}%
\caption{Counts of the classes in the current UPWINS spectral library.}
\label{fig:speciesCount}
\end{figure}

The UPWINS spectral library contains 902 spectra collected with an ASD4 fieldspectrometer from the categories as shown in Figure~\ref{fig:speciesCount}. The library has spectra for 16 vegetation species and 2 soil types. Each spectrum has 2152 bands covering the VNIRSWIR range from 400nm to 2500nm.

\begin{figure}[htb]
\begin{minipage}[b]{1.0\linewidth}
  \centering
	 \centerline{\epsfig{figure=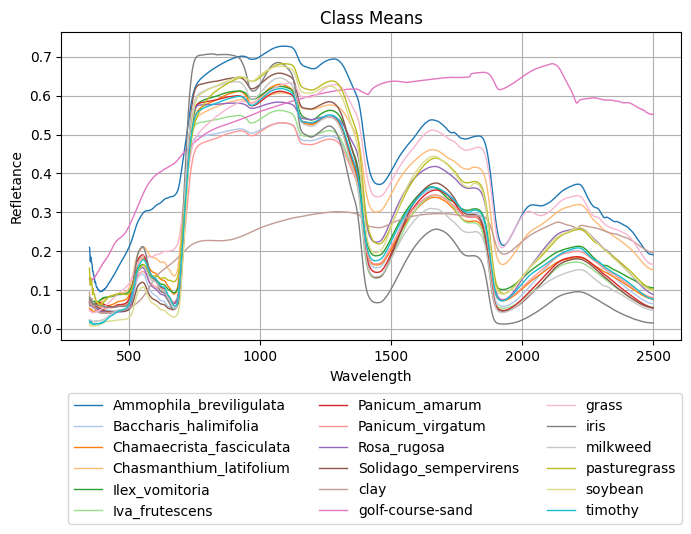, width=8.5cm}}
\end{minipage}%
\caption{The mean spectrum for each class.}
\label{fig:meanSpectra}
\end{figure}

The mean spectrum for each class is shown in Figure~\ref{fig:meanSpectra}.

\section{Methods}
We randomly seperated this dataset into 80\% and 20\% subsets for training and testing. We used the using the LazyClassifier method in the lazypredict. Supervised python package to training each classifier on the training subset, and evaluate the prediction results on the test  subset. The produced accuracy, balanced accuracy, F1 Score, and computational time.

We  trained a neural network (NN) on this partitioned data using TensorFlow.  This NN has a single hidden layer comprised of 128 neurons with ReLu activation functions, followed by a classification layer with with 18 neurons and softmax activation. The NN was trained using sparse categorical crossentropy loss function, 2,000 epochs, and a batch size of 32.

Statistical inference is the process of determining relationships between variables by analyzing parameters in a model, for example examining the coefficients and p-values in a multivariate linear regression to understand the relationship between independent and dependent variables.  We analyze the weights (or coefficients) in the neural network to infer relationships between reflectance values at specific wavelengths and the material classes. 

\section{Results}
The performace metrics for the machine learning model comparison is shown in Table~\ref{table:MLaccuracy}, sorted by accuracy. 
\begin{table}[ht]
\scriptsize
\centering
\begin{tabular}{|l|c|c|c|c|}
\hline
\textbf{Model} & \textbf{Accuracy} & \textbf{Balanced Acc} & \textbf{F1 Score} & \textbf{Time Taken} \\
\hline
LDA & 0.91 & 0.91 & 0.92 & 0.61 \\
RidgeClassifierCV & 0.87 & 0.86 & 0.87 & 0.23 \\
ExtraTreesClassifier & 0.87 & 0.85 & 0.87 & 0.63 \\
BaggingClassifier & 0.86 & 0.84 & 0.86 & 13.11 \\
RandomForestr & 0.86 & 0.84 & 0.86 & 2.07 \\
LGBMClassifier & 0.85 & 0.84 & 0.86 & 50.84 \\
XGBClassifier & 0.85 & 0.83 & 0.85 & 63.16 \\
ExtraTreeC & 0.80 & 0.81 & 0.81 & 0.08 \\
CalibratedCV & 0.82 & 0.80 & 0.82 & 96.91 \\
RidgeClassifier & 0.81 & 0.79 & 0.81 & 0.14 \\
LinearSVC & 0.78 & 0.76 & 0.79 & 20.58 \\
DecisionTree & 0.78 & 0.75 & 0.78 & 1.45 \\
LabelSpreading & 0.71 & 0.69 & 0.74 & 0.16 \\
LabelPropagation & 0.71 & 0.69 & 0.74 & 0.16 \\
LogisticRegression & 0.73 & 0.68 & 0.73 & 0.71 \\
KNearestNeighbors & 0.62 & 0.57 & 0.62 & 0.10 \\
SGDClassifier & 0.58 & 0.51 & 0.59 & 0.79 \\
PassiveAggressive & 0.56 & 0.48 & 0.56 & 1.35 \\
SVC & 0.53 & 0.47 & 0.52 & 0.59 \\
Perceptron & 0.45 & 0.41 & 0.46 & 0.54 \\
GaussianNB & 0.39 & 0.36 & 0.35 & 0.15 \\
QDA & 0.38 & 0.33 & 0.34 & 0.31 \\
BernoulliNB & 0.36 & 0.32 & 0.32 & 0.11 \\
NearestCentroid & 0.36 & 0.30 & 0.35 & 0.08 \\
AdaBoostClassifier & 0.27 & 0.20 & 0.16 & 18.24 \\
\hline
\end{tabular}
\caption{Performance metrics of various classifiers, sorted by accuracy.}
\label{table:MLaccuracy}
\end{table}

\begin{figure}[htb]
\begin{minipage}[b]{1.0\linewidth}
  \centering
	 \centerline{\epsfig{figure=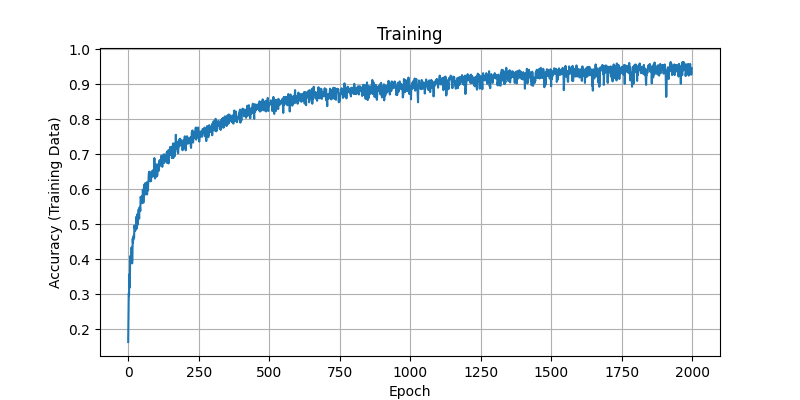, width=8.5cm}}
\end{minipage}%
\caption{The accuracy of the neural network on the training data during the training process.}
\label{fig:trainingHistoryAccuracy}
\end{figure}

The accuracy (on the training data) during training of the NN is shown in Figure~\ref{fig:trainingHistoryAccuracy}.

\begin{figure}[htb]
\begin{minipage}[b]{1.0\linewidth}
  \centering
	 \centerline{\epsfig{figure=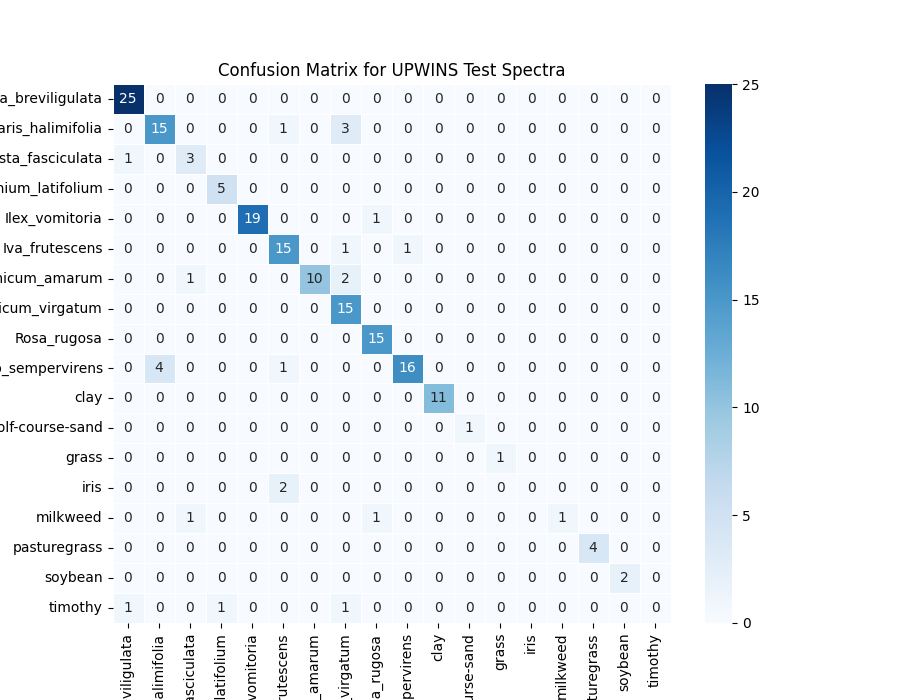, width=8.5cm}}
\end{minipage}%
\caption{The confusion matrix for our neural network classification of the test subset.}
\label{fig:confusionMatrix}
\end{figure}

After training, the NN was applied to the test set resulting in an accuracy of 0.87.  The confusion matrix for prediction on the test set is shown in Figure~\ref{fig:confusionMatrix}.

\begin{figure}[htb]
\begin{minipage}[b]{1.0\linewidth}
  \centering
	 \centerline{\epsfig{figure=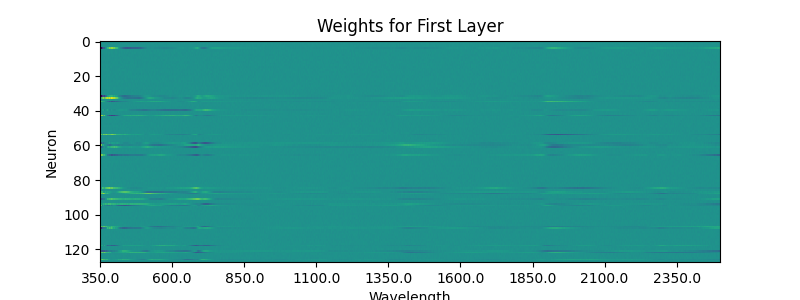, width=8.5cm}}
\end{minipage}%
\caption{The matrix $W^1$ of the weights in layer 1 of the neural network.}
\label{fig:Layer1WeightsImAll}
\end{figure}

The 128x2152 matrix of weights for the first layer are shown in Figure~\ref{fig:Layer1WeightsImAll}, which we refer to as $W^1$. Each row in this matrix is the set of weights (or parameters or coefficients) that are multiplied by the spectral reflectance values within a neuron in the first layer. Specifically, if $W^1_{i,j}$ is the $i,j$-th element in this array, then the output of neuron $i$ in the layer 1 of the network for an input spectrum $matbf{s}=(s_1,..,s_{2152})$ is
\[
f^1_i(\mathbf{s}) = \text{ReLu}( b_i + \sum_{j=1}^{2152} W^1_{i,j} s_i) = \text{ReLu}(b_i + W^1_{i,:}\mathbf{s}),
\]
where $ReLu(x)$ is equal to$x$ if $x>0$ and equal to 0 otherwise.

\begin{figure}[htb]
\begin{minipage}[b]{1.0\linewidth}
  \centering
	 \centerline{\epsfig{figure=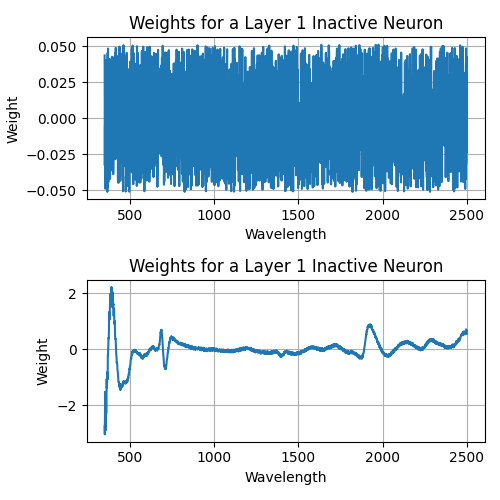, width=8.5cm}}
\end{minipage}%
\caption{Weights for a neuron in layer 1 that is inactive.}
\label{fig:inactiveNeuron}
\end{figure}

An active neuron is a neuron whose weights have been substantively modified during training and that contributes to the NN prediction, while an inactive learning is one that is not contributing. Only some of the neurons show evidence of training, apparent in Figure~\ref{fig:layer1Weights} as rows with variation. The weights for an inactive neuron in layer 1 not showing evidence that its weights were modified from the initial random state as shown in Figure~\ref{fig:inactiveNeuron} (top) along with the weights for an active neuron whose weights were modified from the initial state (bottom). (Our criteria for an active neuron is one in which the standard deviation of the weights is greater than 0.1.) These values are multiplies by the reflectance values in spectra during prediction, and thus play a role similar to regression coefficients but incorporated in nonlinear combinations through subsequent layers. 

\begin{figure}[htb]
\begin{minipage}[b]{1.0\linewidth}
  \centering
	 \centerline{\epsfig{figure=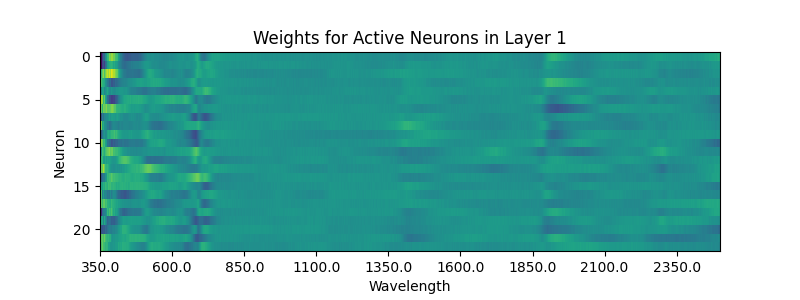, width=8.5cm}}
\end{minipage}%
\caption{Weights in layer 1 of the neural network, showing only the 23 layers with activation.}
\label{fig:Layer1WeightsIm}
\end{figure}

The weights in the plots in Figures~\ref{fig:inactiveNeuron} and~\ref{fig:activeNeuron} are the values in rows 1 and 4 of the matrix of layer 1 weights shown in Figure~\ref{fig:Layer1WeightsImAll}. We can better visualize the weights for neurons used to classification by removing the rows for inactive neurons in the matrix of weights for layer 1 (Figure~\ref{fig:Layer1WeightsImAll}). The resulting array of weights for active neurons is shown in Figure~\ref{fig:Layer1WeightsIm}.

\begin{figure}[htb]
\begin{minipage}[b]{1.0\linewidth}
  \centering
	 \centerline{\epsfig{figure=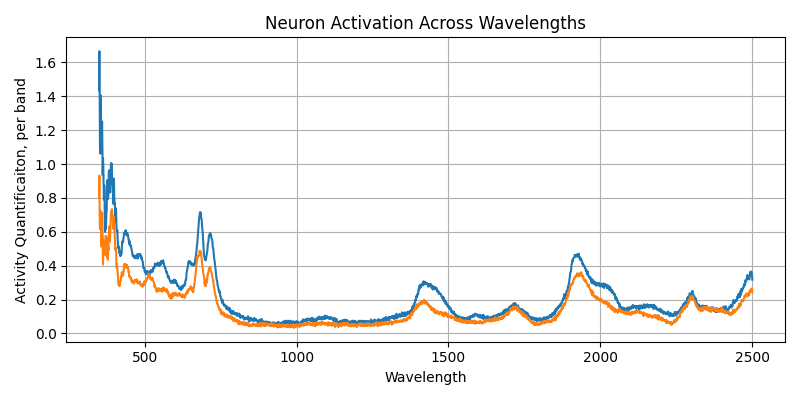, width=8.5cm}}
\end{minipage}%
\caption{Amount of neuron activation as a function of wavelength.}
\label{fig:NeuronActivityPerWL}
\end{figure}

The weights shown in Figure~\ref{fig:Layer1WeightsIm} provide insight in the portions of the spectra most useful for prediction. Each column in the matrix shown corresponds to a spectral band with associated wavelength, and these wavelengths are provided along the x-axis (bottom edge).  Wavelegnths that are useful for prediction are indicated by variation in the values shown in the associated column.  A plot of the mean and standard deviation in neuron activation as a function of wavelength (mean and standard deviation of the weight matrix shown in Figure~\ref{fig:Layer1WeightsIm}) is shown in Figure~\ref{fig:NeuronActivityPerWL}.

From Figure~\ref{fig:NeuronActivityPerWL}, we see that the there is significant activity across the 350nm-750nm rangee, there is activity in the 1400nm-1500nm and 1900nm-200nm ranges, and  isolated activity at approximately 1700nm, 2300nm, and 2500nm.

\begin{figure}[htb]
\begin{minipage}[b]{1.0\linewidth}
  \centering
	 \centerline{\epsfig{figure=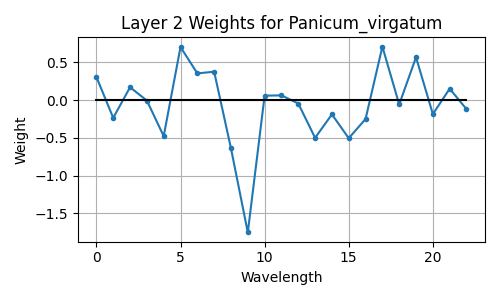, width=8.5cm}}
\end{minipage}%
\caption{The weights in the layer 1 neuron for Panicum virgatum. The $i$-th weight indicates how strongly classifybing Panicum virgatum relies on the $i$-th neuron from layer 1.}
\label{fig:PanVir}
\end{figure}

\begin{figure}[htb]
\begin{minipage}[b]{1.0\linewidth}
  \centering
	 \centerline{\epsfig{figure=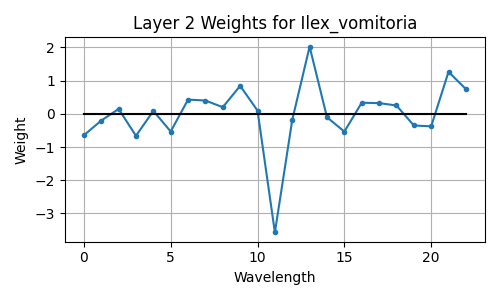, width=8.5cm}}
\end{minipage}%
\caption{The weights in the layer 1 neuron for Panicum virgatum. The $i$-th weight indicates how strongly classifybing Panicum virgatum relies on the $i$-th neuron from layer 1.}
\label{fig:IlexVom}
\end{figure}

To infer associations between wavelengths are specific classes, we consider the weights in layer 2. If we denote the vector of outputs of the neurons in layer 1 by $\mathbf{v}$, then the $i^{th}$ neuron in layer two outputs the following value, 
\[
f^2_i(\mathbf{v}) = \text{Softmax}( b^2_i + \sum_{j=1}^{128} W^2_{i,j} v_i).
\]
where the Softmax function normalizes the outputs of the second layer so that they are all positive and sum to one. For each class in our dataset, there is a corresponding neuron in Layer 2 that outputs the probability that the input spectrum belongs to its class. The weights for this neuron indicate how much declaring that class depends on each of the outputs of neurons in layer 1. Plots of the weights for the layer 2 neurons corresponding to Panicum virgatum and Ilex vomitoria are shown in Figures~\ref{fig:PanVir} and~\ref{fig:IlexVom}.

From the plots in Figures~\ref{PanVir} and~\ref{fig:IlexVom}, we see that classification of a spectrum from Panicum virgatum relies mostly on neuron 9 in layer 1, and classification of Ilex vomitorium relies on neurons 11 and 13 (setting a somewhat arbitrary threshold of 1 on the magnitude of the weights).

\begin{figure}[htb]
\begin{minipage}[b]{1.0\linewidth}
  \centering
	 \centerline{\epsfig{figure=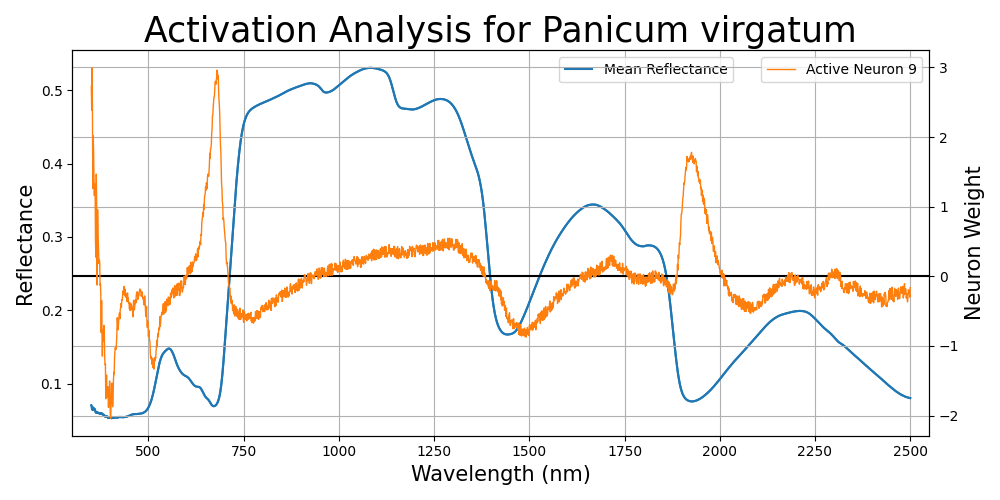, width=8.5cm}}
\end{minipage}%
\caption{Spectral Activation Plot for Panicum virgatum.}
\label{fig:activation_Panicum_virgatum}
\end{figure}

\begin{figure}[htb]
\begin{minipage}[b]{1.0\linewidth}
  \centering
	 \centerline{\epsfig{figure=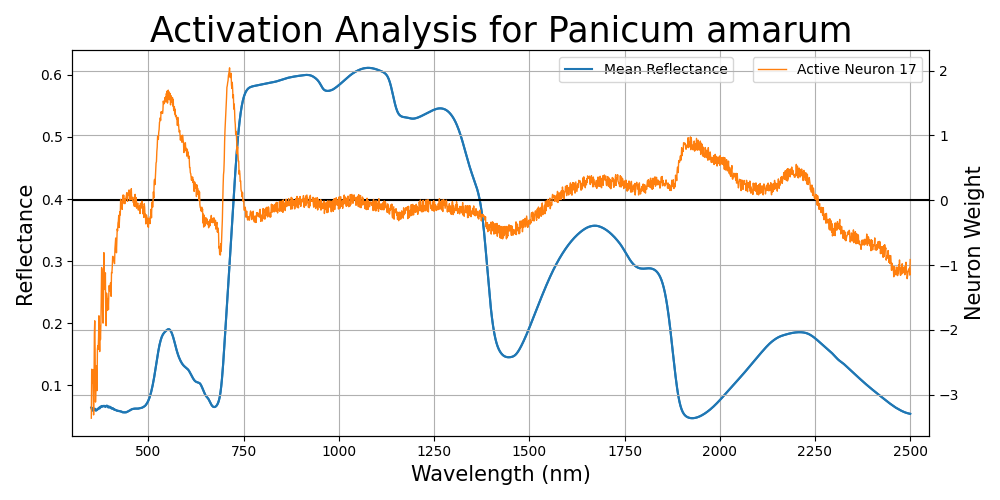, width=8.5cm}}
\end{minipage}%
\caption{Spectral Activation Plot for Panicum amarum.}
\label{fig:activation_Panicum_amarum}
\end{figure}

Very informative plots are shown in Figures~\ref{fig:activation_Panicum_virgatum} through~\ref{fig:activation_Panicum_amarum}. In each plot, we show the mean spectrum for the class (in blue) along with the weight(s) from layer 1 neurons that are heavily utilized for designating this class.  For example, Panicum virgatum heavily utilizes neuron 9 from layer 1, and so Figure~\ref{fig:activation_Panicum_virgatum} shows the mean spectrum along with a plot of $W^2_{7,9}$ (weight 9 shown in Figure~\ref{fig:PanVir}) times the vector of layer 1 weights $W^2_{9,\cdot}$. In each species class shown, the weights that are plotted show how the neural network relies on the different bands for classifying that species. We call these plots Spectral Activation Plots.

\begin{figure}[htb]
\begin{minipage}[b]{1.0\linewidth}
  \centering
	 \centerline{\epsfig{figure=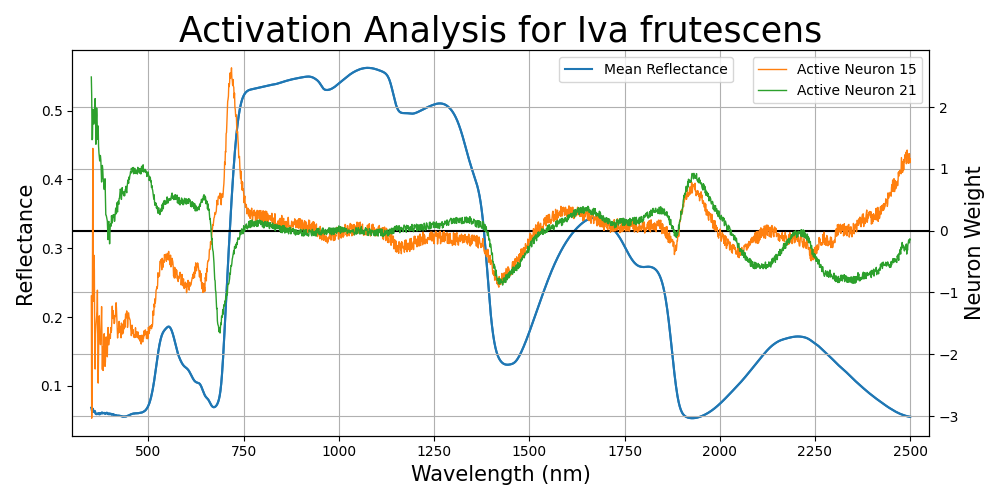, width=8.5cm}}
\end{minipage}%
\caption{Spectral Activation Plot for IVa frutescens.}
\label{fig:activation_Iva_frutescens}
\end{figure}
\begin{figure}[htb]
\begin{minipage}[b]{1.0\linewidth}
  \centering
	 \centerline{\epsfig{figure=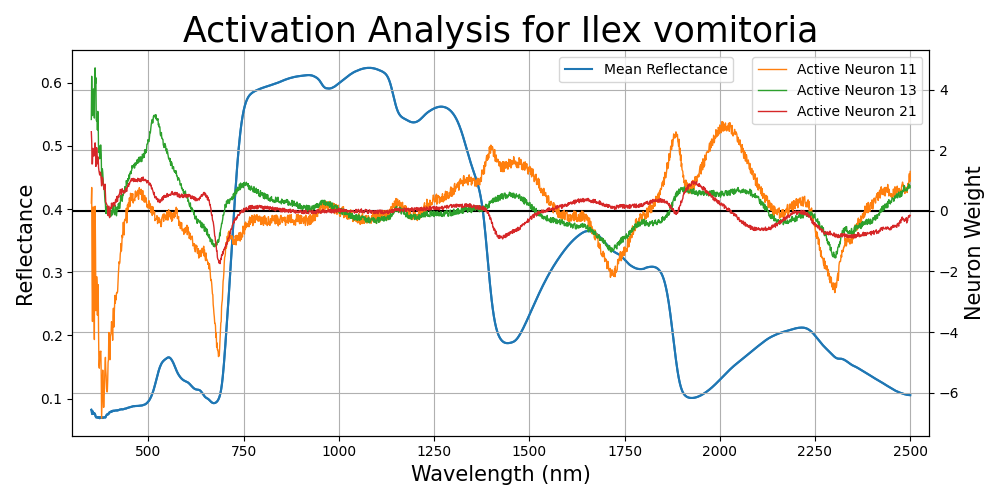, width=8.5cm}}
\end{minipage}%
\caption{Spectral Activation Plot for Ilex vomitoria.}
\label{fig:activation_Ilex_vomitoria}
\end{figure}

\section{Discussion}
The UPWINS spectral library contains ASD spectra for 16 vegetation species and two soil types. We showed that classification on this library can be achieved with 93\% accuracy using an LDA classifier and 87\% accuracy with a neural network.  In each case, the data was split randomly into 80-20 training and testing subsets, with the model trained on the training subset and accuracy is evaluated using the testing subset. Given the number of spectra for each class (particularly the classes with a genus-species name) this suggests that the spectra are sufficiently different for robust classification. 

It is possible that further training or optimization could produce higher accuracy in the LDA or NN model, but maximizing accuracy is not the goal of this paper.  Our goals are: (1) understanding the chemical-physical-biological-functional differences in the classes in the UPWINS library, (1) determining if and how these differences are observable as features in the measured spectra, and (3) how the importance of features for distinguishing between the species based in interpretable model parameters.

This is a promising indication that the values measured in the VNIRSWIR range are sufficient for separation between these species in hyperspectral imagery. We found the most important features are in the 350nm-900nm range, although useful information is present in the NIR and SWIR regions as well.  Detecting these species, and discriminating them from other 'background' vegetation in imagery, would likely require additional spectra collected off the background vegetation types.

We did not evaluate classification or prediction evaluating the additional information provided in the library for most spectra (growth stage, health, stress). There did not seem to be sufficient coverage of these on a per-species case.

We provided analysis of the neural network model, showing that this model not only intrepretable, but in fact that it is useful for statistical inference.  The plots in Figures~\ref{fig:activation_Panicum_virgatum} through~\ref{fig:activation_Panicum_amarum} show which bands are most important for four of the classes considered and how they are weighted.

Panicum vergatim and Panicum amarum grasses of the same genus.  Panicum vergatim, commonly called switchgrass, is a common grass in central North American prarie.  Panicum vergatim grows well in coastal regions, has deep roots, and is used for dune stabilization.  The spectral activation plots for these species (Figures~\ref{fig:activation_Panicum_vergatim} and~\ref{fig:activation_Panicum_amarum}) show that the NN model discriminates between these species using variation in the features from 350nm to 750nm.  These variations likely due to differences in relative amounts of chlorophyll-a and chlorophyll-b. There is rise in the reflectance values for Panicum vergatim at 350nm but a drop in Panicum amarum at 350nm - it would be interesting to determine if this is a stable feature in these species (through further collection under varying conditions as well as statistical analysis of spectra in each species class), and if so what chemical compounds cause this difference.

Iva frutescens is a species of flowering shrub.  It is tolerant of salinity but not tolerant of flooding, so it tends to grow in a narrow band along the edge of salt marshes.  Unlike the activation plots for the Panicum grasses, the neuron weights for Iva frutescens have a distinct upward trend approaching 2500nm. There is also a sharp minima in the weights around 1850nm, which seems to be measuring the broad spectra minima feature across 1800nm-200nm.  This feature is present in all healthy vegetation, and is often attributed to water absorption.  

Ilex vomitoria is a holly that is native to North America. It is the only native plant in the region that contains significant amounts of caffeine. It is drought and shade tolerant. The spectra for this species show a very subtle feature around 2300nm which is not aparent in the class means for other species.  The spectral activation plot shown in Figure~\ref{fig:activation_Ilex_vomitoria} shows that this feature is utilized in classification of this species. The spectral activation plots also show utilization of a feature around 1400nm, evident as a narrow spike the the weights for neuron 11.  There is not an observable feature in the class mean spectrum, but it would be interested to determine if this feature has a physical-chemical meaning.

Our presentation of the spectral activation plots for the neural network model are promising for statistical inference in complex machine learning models.

\bibliographystyle{IEEEbib}
\bibliography{my_refs6}

\end{document}